\documentclass[10pt, a4paper]{article}

\usepackage[]{lrec-coling2024} % this is the new style
\usepackage{multibib}
\usepackage{graphicx}
\usepackage{subfigure}
\usepackage{amsfonts}
\usepackage{bbm}
\usepackage{makecell}
\usepackage{tabularx}
\usepackage{multirow}
\usepackage{booktabs}
\usepackage{soul}
\usepackage{amsmath}
\usepackage{titlesec}
\title{Selective Temporal Knowledge Graph Reasoning}

\name{Zhongni Hou$^{1,2}$, Xiaolong Jin$^{1,2}$$^{*}$, Zixuan Li$^{2}$\sthanks{\ \ Corresponding authors.},  Long Bai$^{2}$, \\ \large{\textbf{Jiafeng Guo$^{1,2}$, Xueqi Cheng$^{1,2}$}}}

\address{$^{1}$School of Computer Science and Technology, University of Chinese Academy of Sciences;\\ $^{2}$Key Laboratory of Network Data Science and Technology,\\ Institute of Computing Technology, Chinese Academy of Sciences. \\
\{houzhongni18z,  jinxiaolong, lizixuan,bailong18b,guojiafeng, cxq\}@ict.ac.cn} 

\abstract{
Temporal Knowledge Graph (TKG), which characterizes temporally evolving facts in the form of (subject, relation, object, timestamp), has attracted much attention recently. TKG reasoning aims to predict future facts based on given historical ones. However, existing TKG reasoning models are unable to abstain from predictions they are uncertain, which will inevitably bring risks in real-world applications. Thus, in this paper, we propose an abstention mechanism for TKG reasoning, which helps the existing models make selective, instead of indiscriminate, predictions. Specifically, we develop a confidence estimator, called Confidence Estimator with History (CEHis), to enable the existing TKG reasoning models to first estimate their confidence in making predictions, and then abstain from those with low confidence.  To do so, CEHis takes two kinds of information into consideration, namely, the certainty of the current prediction and the accuracy of historical predictions. Experiments with representative TKG reasoning models on two benchmark datasets demonstrate the effectiveness of the proposed CEHis.
\\ \newline  
}

\begin{document}

\maketitleabstract
\section{Introduction}

Temporal Knowledge Graphs (TKGs), which store temporally evolving facts in the form of (subject, relation, object, timestamp)~\cite{jin2019recurrent,goel2020diachronic,li2022tirgn}, have emerged as a very active research area over the last few years. Typically, a TKG can be denoted as a sequence of KG snapshots with timestamps, each of which contains all facts at the corresponding timestamp. The TKG reasoning task that aims to, given queries like (query entity, query relation, ?, future timestamp), conduct predictions about these future facts based on historical ones~\cite{ding2021temporal,li2022hismatch}, has recently attracted more and more interest. It has also been increasingly used in various downstream time-sensitive applications, such as emerging event response~\cite{muthiah2015planned,phillips2017using}, policymaking~\cite{deng2020dynamic} and disaster relief~\cite{signorini2011use}.

Although existing models have achieved significant successes in the TKG reasoning task, they still inevitably make incorrect predictions due to the complex temporal dynamics in TKGs. The risk associated with incorrect predictions hinders the more extensive adoption of these models in real-world applications, especially some risk-sensitive applications such as disaster relief~\cite{li2022tirgn} and emergency response~\cite{phillips2017using}. To better facilitate practical applications, it is necessary for the existing TKG reasoning models to have the ability to abstain from making uncertain, even incorrect, predictions.

This kind of ability to abstain from making certain predictions, also known as the selective prediction, has already been studied in the fields of image classification~\cite{whitehead2022reliable,dancette2023improving} and text classification~\cite{kuhn2023semantic}. To make selective predictions, existing studies equip the model with a confidence estimator, which estimates its confidence in the prediction and guides it to abstain from those with low confidence. Those existing studies estimate the confidence of the model mainly based on the final probability distribution of the current prediction. For instance, \citet{geifman2017selective} proposed SoftMax Response (SR) that utilizes the highest probability in the final probability distribution as the model's confidence score. \citet{raina2022answer} and \citet{xin2021art} utilized the entropy of the final probability distribution as the confidence score.

In TKG, besides the model’s confidence in its current prediction, there usually exist some historical predictions that may also help the model decide whether or not to abstain. In fact, there are various historical queries that are relevant to the given query entity and query relation, even the same as the given query. Take the query (ISIS, Attack, ?, 2023-5-13) as an example, there are various related queries in the history, such as those (ISIS, Attack, ?, $t$) occurring before 2023-5-13. If the model can correctly make predictions for most of those related queries, it is very likely to make a correct prediction with high confidence for this query on 2023-5-13. This observation emphasizes the importance of leveraging the accuracy of predictions on historical queries to enhence the models’ confidence in the current prediction.

Motivated by the above issues, this paper studies for the first time the selective prediction setting for the TKG reasoning task. Specifically, we propose a confidence estimator, called Confidence Estimator with History (CEHis), for selective TKG reasoning. CEHis employs a certainty scorer to measure the certainty of the current prediction. It further uses a historical accuracy scorer to model the accuracy of historical predictions, by considering three types of related queries in the history, i.e., query entity related, query relation related, and both query entity and relation related, respectively. As intuitively the impact of the accuracy of the historical predictions on the confidence of the current prediction may decay over time, we employ the Hawkes process~\cite{hawkes2018hawkes} to estimate this impact of long-term and short-term. Finally, CEHis leverages a ranking-based strategy to combine both the certainty score and the historical accuracy score to get the final confidence of the current prediction. Extensive experiments with representative TKG reasoning models on two benchmark datasets demonstrate the effectiveness of CEHis.

In summary, the contributions of this paper are as follows:
\begin{itemize}
\item To facilitate practical applications of TKG reasoning, it studies for the first time the selective TKG reasoning setting;
\item It proposes a simple but effective confidence estimator for this task, which takes both the certainty of the current prediction and the accuracy of historical predictions on related queries into consideration;
\item Experiments on two benchmark datasets demonstrate the necessity of the selective TKG reasoning setting and the superiority of the proposed confidence estimator.
\end{itemize}

\section{Related Work}
\subsection{TKG Reasoning Methods}
There are two different task settings for TKG reasoning, interpolation and extrapolation~\cite{jin2020Renet,park2022evokg,cai2022temporal,messner2022temporal,liu2022tlogic}. The interpolation setting aims to infer missing elements of facts at known timestamps in historical snapshots. In contrast, the extrapolation setting, which this paper focuses on, is to predict future facts.

Under the interpolation setting, HyTE~\cite{dasgupta2018hyte} proposes to conduct TKG reasoning task based on projected-time translation. DE-DistMult~\cite{goel2020diachronic} and DE-SimplE~\cite{goel2020diachronic} both utilize a diachronic embedding to generate entity representations at any given time. However, most interpolated TKG reasoning models perform worse when predicting future temporal facts. Under the extrapolation setting, RE-Net~\cite{jin2020Renet} and REGCN~\cite{li2021temporal} both utilize the recurrent mechanism to capture the complex evolutional patterns among the facts in history. Besides, CyGNet~\cite{zhu2020learning} utilizes a copy-generation mechanism to capture recurrence patterns of temporal facts. Considering that most TKG reasoning methods are black-box, TITer~\cite{sun2021timetraveler} and Cluster~\cite{li2021search} further employ RL to adaptively find history paths, in order to provide interpretations for a specific prediction. More recently, TiRGN~\cite{li2022tirgn} and HisMatch~\cite{li2022hismatch} both design multi-encoders to model different characteristics of historical facts. CENET~\cite{xu2023temporal} further utilizes contrastive learning to identify significant entities from both historical and non-historical dependency. All these methods are encouraged to make predictions even wrong, leading to uncontrollable risks. Different from all these methods, we focus on making selective, instead of indiscriminate, predictions to control the risk of TKG reasoning in this work.

\subsection{Selective Prediction}
The selective prediction setting gives a model an option to abstain from generating certain predictions, which has been explored in different scenarios~\cite{bartlett2008classification,grandvalet2008support,gal2016dropout,cortes2016boosting}. 
When conducting selective prediction, a typical technique is to set a threshold over a confidence score derived from a pre-trained Neural Network. In 2017, \citet{geifman2017selective} proposed to selectively output using the well-known SR and MC-Dropout~\cite{gal2016dropout} as selection strategies.
Recently, SelectiveNet~\cite{geifman2019selectivenet} calculates a confidence score via an additional selection head to determine whether to abstain or not. Similarly, Deep Gamblers~\cite{liu2019deep} and SAT~\cite{huang2020self} introduce an extra abstention class, the corresponding logit of this class determines whether a query is selected to predict or not. More recently, ~\citet{feng2022towards} achieves better results via using the classification scores outputted by selective models with architectural change. Most previous work on selective prediction is mainly applied to CV tasks and the NLP field and mainly focuses on the probability of the current prediction. To the best of our knowledge, we are the first to utilize the characteristics of TKGs and apply selective prediction to the TKG reasoning scenario.

\section{Problem Formulation}
\subsection{Formulation of TKG Reasoning}
A TKG can be formalized as a sequence of KGs with timestamps, i.e., $\{\mathcal{G}_1, \mathcal{G}_2, ..., \mathcal{G}_t, ...\}$. The KG at timestamp $t$ can be denoted as
$\mathcal{G}_t=\{\mathcal{V},\mathcal{R},\mathcal{E}_t\}$, where $\mathcal{V}$, $\mathcal{R}$, $\mathcal{E}_t$ are the sets of entities, relations, and temporal facts occurring at timestamp $t$, respectively. Each fact in $\mathcal{E}_t$ is denoted as $(s,r,o,t)$, where $s,o\in \mathcal{V}$ are the subject and object entities involved in this fact, $r \in \mathcal{R}$ is the relation between $s$ and $o$. The TKG reasoning task aims to predict future facts based on given historical facts, which can be divided into two subtasks, namely, entity reasoning and relation reasoning. The former aims to predict the object (or subject) entity for a given query $q=(s_q,r_q,?,t_q)$ (or $q=(?,r_q, o_q,t_q)$) based on the corresponding history before $t_q$, i.e., $\mathcal{G}_q=\{\mathcal{G}_1,\mathcal{G}_2,...,\mathcal{G}_{t_q-1}\}$. The latter aims to predict the relation for a given query $q=(s_q,?,o_q,t_q)$ based on $\mathcal{G}_q$.

\begin{figure*}[t]
  \centering  
   \includegraphics[width=\linewidth]{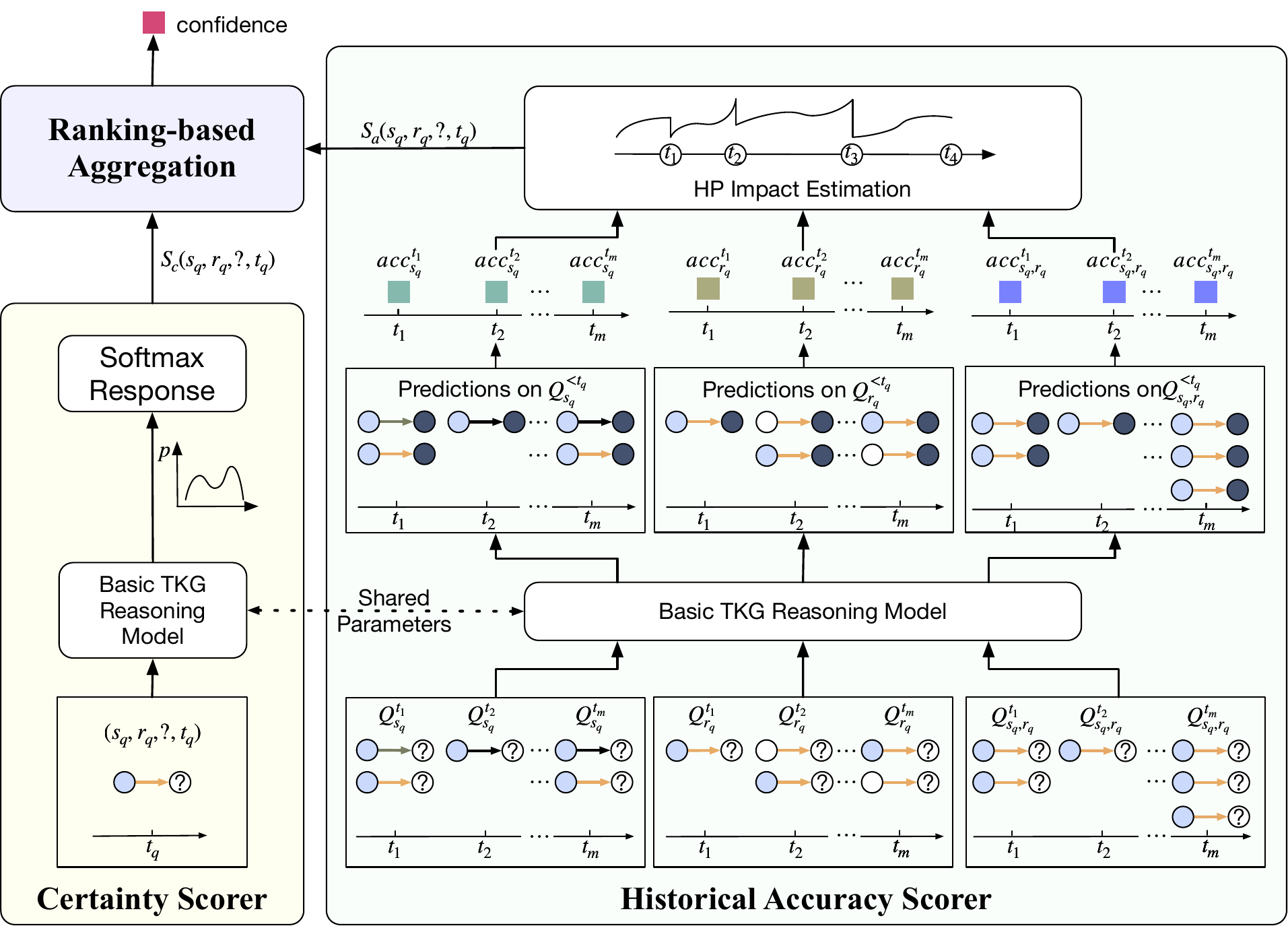}
   \caption{An illustrative diagram of the proposed confidence estimator, CEHis, for selective entity reasoning. For the sake of brevity, the corresponding history $\mathcal{G}_q$ paired with each query $q$ is not explicitly given.}
   \label{fig:model}
   % \vspace{-4mm}
  \end{figure*}

\subsection{Formulation of Selective TKG Reasoning}  
Typically, a selective TKG reasoning model $f_s$ consists of three parts: a basic TKG reasoning model $f$, a confidence estimator $g$ for evaluating the confidence of each prediction, and a threshold $\gamma$ to determine if a prediction should be abstained based on the corresponding confidence. 
When a prediction is trustable, $f_s$ outputs a ranked entity list generated by $f$, otherwise an empty list, i.e., $\emptyset$.

\textbf{The basic TKG reasoning model.} Given an input $x=(q,\mathcal{G}_q)\in \mathcal{X}$, where $q$ denotes the query, $\mathcal{G}_q$ represents the corresponding history and $\mathcal{X}$ is the input space, the model $f$ calculates the probability of the candidate $y\in \mathcal{Y}$ to be the correct answer, i.e., $p(y|x)$, where $\mathcal{Y}$ is the candidate answer space. Note that, for entity reasoning, $\mathcal{Y}$ is the set of entities $\mathcal{V}$, whilst for relation reasoning, $\mathcal{Y}$ is the set of relations $\mathcal{R}$. Finally, $f$ outputs a list of all candidates, ranked in descending order according to their corresponding prediction probability.  

\textbf{The confidence estimator.} The confidence estimator $g(x)$ is a positive real-valued function, which evaluates how confident the model $f(x)$ is on its prediction for a given input $x$. Ideally, $g$ should obtain high values when $f$ makes correct predictions, and otherwise, low values. 

\textbf{The threshold.} Selective prediction seeks to make the trade-off between making correct predictions with high-confidence and abstaining from making low-confidence predictions to control potential risks. Therefore, $f_s$ is equipped with a threshold $\gamma$ to determine whether or not a prediction made by $f$ should be trusted, and further control the overall level of abstention.

Above all, the selective TKG reasoning can be formulated as follows: 

\begin{equation}
% \vspace{-2mm}
f_s(x)=
\begin{cases}
f(x),& if\,\, g(x)>\gamma,\\ 
\emptyset, & if\,\, g(x)\le\gamma.\\ 
\end{cases}
\end{equation}

Obviously, the key in selective TKG reasoning is to assess the confidence of a prediction. Therefore, in this paper we develop a universal confidence estimator that can be readily integrated with existing TKG reasoning models, to estimate their predictions (see Section~\ref{cehis}).
\subsection{Evaluation Metrics of Selective TKG Reasoning}
\label{evaluation}
For selective prediction, \textit{coverage}, \textit{risk} and \textit{effective reliability} are three widely adopted metrics~\cite{geifman2017selective,whitehead2022reliable}. In what follows, we formulate these metrics for the selective TKG reasoning task. Let $\mathcal{D}=\{(x_i,y_i^*)\}_{i=1}^{|\mathcal{D}|}\subseteq \mathcal{X}\times \mathcal{Y}$ be a set of inputs and their corresponding ground truth answers, i.e., $y_i^*$ is the ground truth corresponding to the input $x_i=(q_i,\mathcal{G}_{q_i})$.

\textbf{Coverage and Risk.}
The \textit{coverage}~\cite{geifman2017selective} of $f_s$ on $\mathcal{D}$ is the proportion of predictions that are not abstained on the entire dataset, namely,
\begin{equation}
% \vspace{-2mm}
  {C}(f_s,\mathcal{D})=\frac{1}{|\mathcal{D}|}\sum_{(x_i,y_i^*)\in \mathcal{D}}\mathbbm{1}[g(x_i)>\gamma],
\end{equation}
where $\mathbbm{1}(P)$ is the indicator function that obtains 1 if $P$ is true, otherwise 0. 

As for the \textit{risk} metric, let’s first define that on a specific prediction. Obviously, for a given query, the higher the position of the corresponding ground truth in the final ranked list of candidate answers, the lower the risk of the model. Therefore, for a given input $x=(q,\mathcal{G}_q)$ and its corresponding ground truth $y^*$, its corresponding \textit{risk} is formulated in this paper as follows: 
\begin{equation}
risk_{x}(y^*)= \alpha(1-\frac{1}{R_{x}(y^*)}),
\end{equation}
where $\alpha$ is the risk parameter and $\alpha\ge 1$. 
Accordingly, the risk of $f_s$ on $\mathcal{D}$ can thus be defined as:
\begin{equation}
{R}(f_s,\mathcal{D})=\frac{\sum_{(x_i,y_i^*)\in \mathcal{D}}risk_{x_i}(y_i^*)\cdot \mathbbm{1}[g(x_i)>\!\gamma]}{{C}(f_s,\mathcal{D})}.
\end{equation}
Based on \textit{coverage} and \textit{risk}, the overall performance of $f_s$ on $\mathcal{D}$ can be measured by the Area Under risk-coverage Curve (AUC), which plots risk against coverage ~\cite{geifman2017selective}. And given a threshold $\gamma$, the lower the value of AUC, the better the performance as it represents lower average risk.

\textbf{Effective Reliability.} 
This metric is first proposed in~\citet{whitehead2022reliable} to measure the effectiveness of a selective Visual Question Answering (VQA) model via assigning a penalty to the model when wrong predictions are outputted. Unlike VQA, the size of the candidate answer space in TKGs is relatively large, which increases the difficulty of making precise predictions. Motivated by this, in selective TKG reasoning, for a given input $x$ and the corresponding ranked candidate list generated by the basic TKG reasoning model, we assign a penalty to $f_s$ if the ground truth $y^*$ is not in the top $N$ positions, a reward to $f_s$ if the ground truth $y^*$ is in the top $N$ positions, and no reward if $f_s$ abstains from making a prediction. Here, $N$ can be seen as the tolerance of the model. Formally, the \textit{effective reliability} of $f_s$ on a given input $x$ in TKGs can be defined by:
\begin{equation}
% \vspace{-2mm}
  \!\phi_{c,N}(f_s,x)\!\!=\!
  \!\!\begin{cases}
    \frac{1}{R_x(y^*)},\!\!\!\!\!& if\,\, g(x)>\gamma\;\&\;R_x(y^*)\!\!<N,\\
    -c, \!\!\!\!\!&if\,\, g(x)>\gamma\;\&\;R_x(y^*)\!\!\ge N,\\
    0, \!\!\!\!\!&if\,\, g(x)\le \gamma,
    \end{cases}
\end{equation}
where $c$ is the penalty, and $\frac{1}{R_x(y^*)}$ denotes the corresponding reward. The \textit{effective reliability} of $f_s$ on the entire dataset $\mathcal{D}$ can thus be obtained as 
% {\small{}{}{}{} 
\begin{equation}
% \vspace{-2mm}
\Phi_{c,N}(f_s,\mathcal{D})=\frac{1}{|\mathcal{D}|}\sum_{(x_i,y_i^*)\in \mathcal{D}}\phi_{c,N}(f_s,x_i). 
\end{equation}
% }{\small\par}

\noindent Obviously, the higher the effective reliability, the better the performance of the TKG reasoning model.
\section{Confidence Estimator with History}
\label{cehis}
This section presents the proposed confidence estimator, i.e., CEHis, for selective TKG reasoning. As illustrated in Figure~\ref{fig:model}, CEHis mainly consists of two components, i.e., a certainty scorer and a historical accuracy scorer, to estimate the confidence score of a prediction generated by the basic TKG reasoning model $f$. The former measures the model's certainty of the current prediction, and the latter qualifies the impact of the accuracy of historical predictions on whether to abstain from the current prediction. Furthermore, it aggregates the above two kinds of information using a ranking-based strategy to determine the final confidence score, and subsequently, abstains from predictions with low confidence. 
In the following, we take selective entity reasoning as an example to illustrate how CEHis estimates the confidence of a prediction.
\subsection{The Certainty Scorer}
Given an input $x=(q,\mathcal{G}_q)$, existing TKG reasoning models usually utilize the softmax activation and finally output the probability of each entity to be the correct answer. Typically, the correct predictions tend to have greater maximum probabilities than those incorrect ones~\cite{hendrycks2016baseline}. This characteristic can be utilized to estimate the level of certainty in the current prediction. As a result, in this paper, we adopt the widely used SR to measure the model's certainty of the current prediction, i.e., $S_{c}(x)$, as follows:

\begin{equation}
% \vspace{-2mm}
  S_{c}(x)=\mathop{max}\limits_{o\in \mathcal{V}}p(o|x),
 \end{equation}

\noindent where $p({o|x})$ is the corresponding probability of the entity $o$ to be the correct answer.

\subsection{The Historical Accuracy Scorer}
As mentioned above, a selective TKG reasoning model $f_s$ abstains from the incorrect predictions made by the basic TKG reasoning model $f$ to control the potential risks. However, whether a prediction on a query is correct or not is unknown, as the corresponding fact has not yet occurred. Typically, there are various historical queries that are relevant to the query entity and the query relation. The accuracy of the historical predictions regarding these related queries can reflect the difficulty of the current query, and further can serve as an indicator of whether to trust $f$'s prediction on the current query. Motivated by this, the historical accuracy scorer estimates the accuracy of $f$'s predictions based on the accuracy of historical predictions on three kinds of related queries, i.e., the subject related ones $Q_{s_q}^{<t_q}$, the relation related ones $Q_{r_q}^{<t_q}$, as well as the subject and relation related ones $Q_{s_q,r_q}^{<t_q}$. Considering that the accuracy of recent historical predictions is more important than older ones, the historical accuracy scorer utilizes the Hawkes process to model the time-varying impact of these historical predictions on the confidence of the current prediction, and finally calculates the historical accuracy score of the current prediction.

Specifically, the subject related queries at timestamp $t_i$ consist of queries with $s_q$ as the subject, and are denoted as $\mathcal{Q}_{s_q}^{t_i}=\{(s_q,-,?,t_i)\}_{i=1}^{K_1}$. Here, ``$-$'' means that the corresponding element can be any relations, $K_1$ is the size of $\mathcal{Q}_{s_q}^{t_i}$, and $t_i<t_q$. Similarly, at timestamp $t_i$, the relation related queries are denoted as $\mathcal{Q}_{r_q}^{t_i}=\{(-,r_q,?,t_i)\}_{i=1}^{K_2}$, the subject and relation related queries are denoted as $\mathcal{Q}_{s_q,r_q}^{t_i}=\{(s_q,r_q,?,t_i)\}_{i=1}^{K_3}$. Taking $\mathcal{Q}_{s_q}^{<t_q}$ as an example, we illustrate how to calculate the impact of the historical predictions on the confidence of the current prediction, i.e., ${S}_{s_q}^{<t_q}$. Given a query $\tilde{q}=(s_q,r_i,?,t_i)\in \mathcal{Q}_{s_q}^{t_i}$, CEHis first utilizes the basic TKG reasoning model $f$ to make a prediction based on the corresponding history $\mathcal{G}_{\tilde{q}}$, and calculates the position of the ground truth ${o}_{\tilde{q}}^*$, i.e., 
$R_{\tilde{q},\mathcal{G}_{\tilde{q}}}(o_{\tilde{q}}^*)$, over the ranked entity list generated by $f$. At timestamp $t_i$, the accuracy of historical predictions on the subject related queries, i.e., $acc_{s_q}^{t_i}$, is calculated by:
 \begin{equation}
 % \vspace{-2mm}
  acc_{s_q}^{t_i}=\frac{1}{|\mathcal{Q}_{s_q}^{t_i}|}\sum_{\tilde{q}\in \mathcal{Q}_{s_q}^{t_i}}\frac{1}{R_{\tilde{q},\mathcal{G}_{\tilde{q}}}(o_{\tilde{q}}^*)}.
 \end{equation} 
After processing all historical snapshots, we can obtain a prediction accuracy sequence of $\mathcal{Q}_{s_q}^{<t_q}$, i.e., 
$Acc_{s_q}^{<t_q}=\{acc_{s_q}^{t_1},...,acc_{s_q}^{t_i},...,acc_{s_q}^{t_m}\}$, where $t_1<...<t_i<...<t_m<t_q$.

To precisely estimate whether the prediction made by $f$ is correct or not, CEHis considers the impact of both long-term and short-term accuracy regarding historical predictions. Typically, the accuracy of the historical predictions on a recent timestamp is more important than that on an earlier timestamp. As a result, it is necessary to take the time information into consideration. Motivated by this, CEHis utilizes the Hawkes process (``HP Impact Estimation'' in Figure~\ref{fig:model};~\citet{cox1980point,laub2015hawkes,zuo2020transformer}) to estimate the impact of the accuracy of historical predictions on whether to abstain or not as follows:

\begin{equation}
% \vspace{-2mm}
S_{s_q}^{<t_q}=\!\!\mu_{s_q}^{<t_q}+\!\sum_{h=0}^{l-1}\!k(t_q-t_{m-h})acc_{s_q}^{t_{m-h}},
  \label{eq:Hawkes}
\end{equation}
where $\mu_{s_q}^{<t_q}$ represents the base prediction accuracy (the long-term accuracy) of the subject related queries, which is calculated by the mean of $Acc_{s_q}^{<t_q}$; $l$ is a length hyperparameter which is used to truncate the prediction accuracy sequence $Acc_{s_q}^{<t_q}$. 
$k(\cdot)$ is a predefined decaying function, calculating the decaying impact of the historical accuracy:
\begin{equation}
% \vspace{-2mm}
    k(t_q-t_h)=exp(-\delta(t_q-t_h)),
\end{equation}
\noindent where $\delta\ge 0$ denotes the decay rate. Besides the above absolute time interval, we can also choose the relative time order information~\cite{zhang2022dynamic}, which can be seen as a normalization, to calculate the decaying impact of the historical accuracy. Obviously, the cumulative term describes that the historical prediction accuracy of the latest timestamps (the short-term accuracy) has a positive contribution to whether to trust $f$'s current prediction.
Similarly, at the query timestamp $t_q$, we can derive the impact of the accuracy of historical predictions regarding $\mathcal{Q}_{r_q}^{<t_q}$ and $\mathcal{Q}_{s_q,r_q}^{<t_q}$, namely, $S_{r_q}^{<t_q}$ and $S_{s_q,r_q}^{<t_q}$, respectively.

For a given input $x=(q,\mathcal{G}_q)$, the final historical accuracy score is calculated as:
% {\small{}{}{}{} 
\begin{equation}
% \vspace{-2mm}
  S_{a}(x)=S_{s_q}^{<t_q}+S_{r_q}^{<t_q}+S_{s_q,r_q}^{<t_q}.
\end{equation}
% }{\small\par}

\begin{table}[t]
  \centering
\renewcommand\arraystretch{1.1}
  \resizebox{0.95\linewidth}{!}{
  \begin{tabular}{cccccccc}
  \toprule
  Datasets &\#Train &\#Valid &\#Test &\#Ent &\#Rel\\
  \midrule
  ICEWS14  &74,845     &8,514  &73,71  &6,869  &230\\
  ICEWS18   &373,018    &45,995 &49,545 &23,033 &256\\
  \bottomrule
  \end{tabular}
  }
  \caption{Statistics of the datasets.}
  \label{table:dataset}
 % \vspace{-5mm}
  \end{table}
\subsection{The Ranking-based Aggregation}
To calculate the final confidence scores of predictions made by $f$,
we need to aggregate two kinds of scores outputted by the above two scorers. However, the certainty score and the historical accuracy score are calculated based on different information and are on different scales. As a result, they cannot be directly combined using absolute values. To this end, CEHis utilizes a ranking-based aggregation strategy, which first ranks queries according to the above two different scores, and then calculates the final confidence score of two based on the results of the two rankings.

More specifically, let $\mathcal{Q}_{un}=\{(q_i,\mathcal{G}_{q_i})\}_{i=1}^{|\mathcal{Q}_{un}|}$ denote a set of query-history pairs that require predictions with unknown distribution. Given an input $x=(q,\mathcal{G}_q)\in \mathcal{Q}_{un}$, we can derive the model's certainty rank, i.e., 
$R_{c,\mathcal{Q}_{un}}(x)$, and the historical accuracy rank, i.e., $R_{a,\mathcal{Q}_{un}}(x)$, of its prediction among $\mathcal{Q}_{un}$:

\begin{equation}
R_{c,\mathcal{Q}_{un}}(x)=\sum_{x_i\in \mathcal{Q}_{un}}\mathbbm{1}(S_c(x_i)>S_c(x)),
\end{equation}
\begin{equation}
R_{a,\mathcal{Q}_{un}}(x)=\sum_{x_i\in \mathcal{Q}_{un}}\mathbbm{1}(S_a(x_i)>S_a(x)).
\end{equation}
The total confidence score generated by our confidence estimator $g$ is defined as:
\begin{equation}
  g(x)=\beta R_{c,\mathcal{Q}_{un}}(x)+(1-\beta) R_{a,\mathcal{Q}_{un}}(x),
\end{equation}
where $\beta$ is an aggregation weight that can be set using the validation dataset.

\begin{table*}[t]
\centering
\resizebox{0.9\linewidth}{!}{
\begin{tabular}{cccccccccc}
\toprule

\multicolumn{1}{c|}{} 
& \multicolumn{1}{c|}{} & \multicolumn{4}{c|}{\multirow{1}{*}{ICEWS14}}& \multicolumn{4}{c}{\multirow{1}{*}{ICEWS18}} \\ 
\cmidrule(r){3-6}  \cmidrule(r){7-10} 

\multicolumn{1}{c|}{Model} & \multicolumn{1}{c|}{Confidence Estimator}& \multicolumn{3}{c|}{\multirow{1}{*}{coverage}}&\multicolumn{1}{c|}{\multirow{2}{*}{AUC}}& \multicolumn{3}{c|}{\multirow{1}{*}{coverage}}&\multicolumn{1}{c}{\multirow{2}{*}{AUC}}\\

\cmidrule(r){3-5}  \cmidrule(r){7-9} 
\multicolumn{1}{c|}{} & \multicolumn{1}{c|}{}& risk=0.1 &risk=0.3&risk=0.5&\multicolumn{1}{|c|}{}&risk=0.1 &risk=0.3&risk=0.5&\multicolumn{1}{|c}{} \\  

\midrule
\multicolumn{1}{c|}
{\multirow{6}{*}{RENET}}& \multicolumn{1}{c|}{EN} &1.81&17.68&61.90&\multicolumn{1}{|c|}{43.34}&0.01&4.99&22.64&\multicolumn{1}{|c}{56.82}\\
\multicolumn{1}{c|}{}&\multicolumn{1}{c|}{SR}     &1.46&14.62&62.15&\multicolumn{1}{|c|}{43.01}&0.01&4.16&22.26&\multicolumn{1}{|c}{56.76}\\
\multicolumn{1}{c|}{}& \multicolumn{1}{c|}{SNR}   &1.24 &17.16&62.76&\multicolumn{1}{|c|}{42.84}&\textbf{0.22}&3.05&21.99&\multicolumn{1}{|c}{56.95}\\
\multicolumn{1}{c|}{}& \multicolumn{1}{c|}{SATR}  &1.37 &16.70&62.84&\multicolumn{1}{|c|}{42.76}&0.02&4.04&22.44&\multicolumn{1}{|c}{57.03}\\
\multicolumn{1}{c|}{}& \multicolumn{1}{c|}{SE}     &1.78 &15.32&60.19&\multicolumn{1}{|c|}{43.65}&0.02&4.07&21.25&\multicolumn{1}{|c}{57.38}\\
\multicolumn{1}{c|}{} & \multicolumn{1}{c|}{CEHis}  &\textbf{3.48}&\textbf{24.97}&\textbf{63.75}&\multicolumn{1}{|c|}{\textbf{40.94}}&0.08&\textbf{7.43}&\textbf{27.05}&\multicolumn{1}{|c}{\textbf{55.14}}\\
\midrule
\multicolumn{1}{c|}
{\multirow{6}{*}{REGCN}}& \multicolumn{1}{c|}{EN}&2.00 &23.11&69.38&\multicolumn{1}{|c|}{40.16}&0.24&6.76&30.51&\multicolumn{1}{|c}{53.20}\\
\multicolumn{1}{c|}{} &\multicolumn{1}{c|}{SR}     &1.98 &23.33&68.33&\multicolumn{1}{|c|}{40.26}&0.61&7.09&31.88&\multicolumn{1}{|c}{52.76}\\
\multicolumn{1}{c|}{}& \multicolumn{1}{c|}{SNR}   &1.57 &24.87&68.92&\multicolumn{1}{|c|}{40.04}&0.52&6.09&31.05&\multicolumn{1}{|c}{53.37}\\
\multicolumn{1}{c|}{}& \multicolumn{1}{c|}{SATR}  &3.71 &25.20&68.93&\multicolumn{1}{|c|}{39.67}&0.28&6.33&29.58&\multicolumn{1}{|c}{53.86}\\
\multicolumn{1}{c|}{}& \multicolumn{1}{c|}{SE}     &\textbf{5.14} &23.99&61.73&\multicolumn{1}{|c|}{42.49}&0.13&4.96&26.39&\multicolumn{1}{|c}{54.84}\\
\multicolumn{1}{c|}{} & \multicolumn{1}{c|}{CEHis}  &4.86 &\textbf{27.51}&\textbf{69.49}&\multicolumn{1}{|c|}{\textbf{38.76}}&\textbf{0.75}&\textbf{8.90}&\textbf{33.58}&\multicolumn{1}{|c}{\textbf{51.92}}\\
\midrule
\multicolumn{1}{c|}
{\multirow{6}{*}{TiRGN}} & \multicolumn{1}{c|}{EN}&3.79 &26.18&75.34&\multicolumn{1}{|c|}{38.41} &0.00&5.96&29.76&\multicolumn{1}{|c}{52.82}\\
\multicolumn{1}{c|}{} &\multicolumn{1}{c|}{SR}     &4.15 &29.83&76.56&\multicolumn{1}{|c|}{37.41}    &0.00&6.72&31.81&\multicolumn{1}{|c}{52.27}\\
\multicolumn{1}{c|}{}& \multicolumn{1}{c|}{SNR}   &4.23 &28.92&72.70&\multicolumn{1}{|c|}{37.86} &0.17&7.52&33.40&\multicolumn{1}{|c}{51.85}\\
\multicolumn{1}{c|}{}& \multicolumn{1}{c|}{SATR}  &4.30 &26.84&72.56&\multicolumn{1}{|c|}{38.36} &0.06&6.75&30.47&\multicolumn{1}{|c}{52.76}\\
\multicolumn{1}{c|}{}& \multicolumn{1}{c|}{SE}    &{5.14} &23.99&72.19&\multicolumn{1}{|c|}{39.42} &0.01&5.79&30.23&\multicolumn{1}{|c}{52.97}\\
\multicolumn{1}{c|}{} & \multicolumn{1}{c|}{CEHis}  &\textbf{5.19} &\textbf{30.32}&\textbf{76.77}&\multicolumn{1}{|c|}{\textbf{36.97}}   &\textbf{0.26}&\textbf{8.16}&\textbf{34.60}&\multicolumn{1}{|c}{\textbf{51.50}}\\
\bottomrule
\end{tabular}}
\caption{Risk-coverage metrics results and AUC results of the selective entity reasoning task. The risk ${R}$ is set to 0.1, 0.3, and 0.5. }
\label{table:mainpred1}
% \vspace{-5mm}
\end{table*}

\section{Experiments}
\subsection{Experiment Setting}

\textbf{Datasets.} We conduct experiments on two widely used TKGs, namely, ICEWS14~\cite{li2022hismatch} and ICEWS18~\cite{li2022tirgn}. They both, with the time granularity of 24 hours, are subsets of facts in the Integrated Crisis Early Warning System (ICEWS). Specifically, ICEWS14 and ICEWS18 contain facts that took place between 2014 and 2018, respectively. The statistics of these datasets is presented in Table~\ref{table:dataset}.

\textbf{Basic TKG Reasoning Models.}
As aforementioned, the proposed confidence estimator, CEHis, can be applied to a variety of existing TKG reasoning models. Here, we take the following three representative TKG reasoning models with varying architectures and performance, as the basic models: {RENET}~\cite{jin2020Renet}, which employs the recurrent GCN to model the histories of queries; {REGCN}~\cite{li2021temporal}, which stacks GCN layers to mine evolutional patterns of each entity among facts occurring at the latest timestamps; {TiRGN}~\cite{li2022tirgn}, which utilizes an additional encoder to capture the structure dependency among repetitive history and is the most relevant model to our confidence estimator.

\textbf{Baselines.}
Since the selective TKG reasoning task has not been explored before, we adopt a few representative confidence estimators used in other tasks as our baselines, including {SR}~\cite{geifman2017selective}, {entropy} (EN;~\citet{geifman2017selective}), {SelectiveNet+SR} (SNR;~\citet{feng2022towards}, {SAT+SR} (SATR;~\citet{feng2022towards}) and {Deep Sub-Ensembles} (SE;~\citet{valdenegro2023sub}). Specifically, SNR and SATR are built upon SelectiveNet~\cite{geifman2019selectivenet} and SAT~\cite{huang2020self}, respectively. They both train a more general classifier by changing the model architectures, and utilize the maximum probability of the classifier itself to conduct selective prediction. SE ensembles only a selection of the model's layers that are close to the output, and estimates the confidence using the predictive uncertainty.

\textbf{Implementation Details.}
We set both the decay rate $\delta$ and the aggregation weight $\beta$ within [0,1], and $\alpha$ to be one. The long-term accuracy of historical predictions is calculated based on the latest 10 timestamps, and the short-term accuracy is calculated based on the latest 3 timestamps. All basic TKG reasoning models are trained using their reported optimal parameters.

\subsection{Experimental Results}
\subsubsection{Results on Selective Entity Reasoning}
\begin{figure} 
    \centering
  \subfigure[RENET, N=5.] {
  \includegraphics[width=0.465\linewidth]{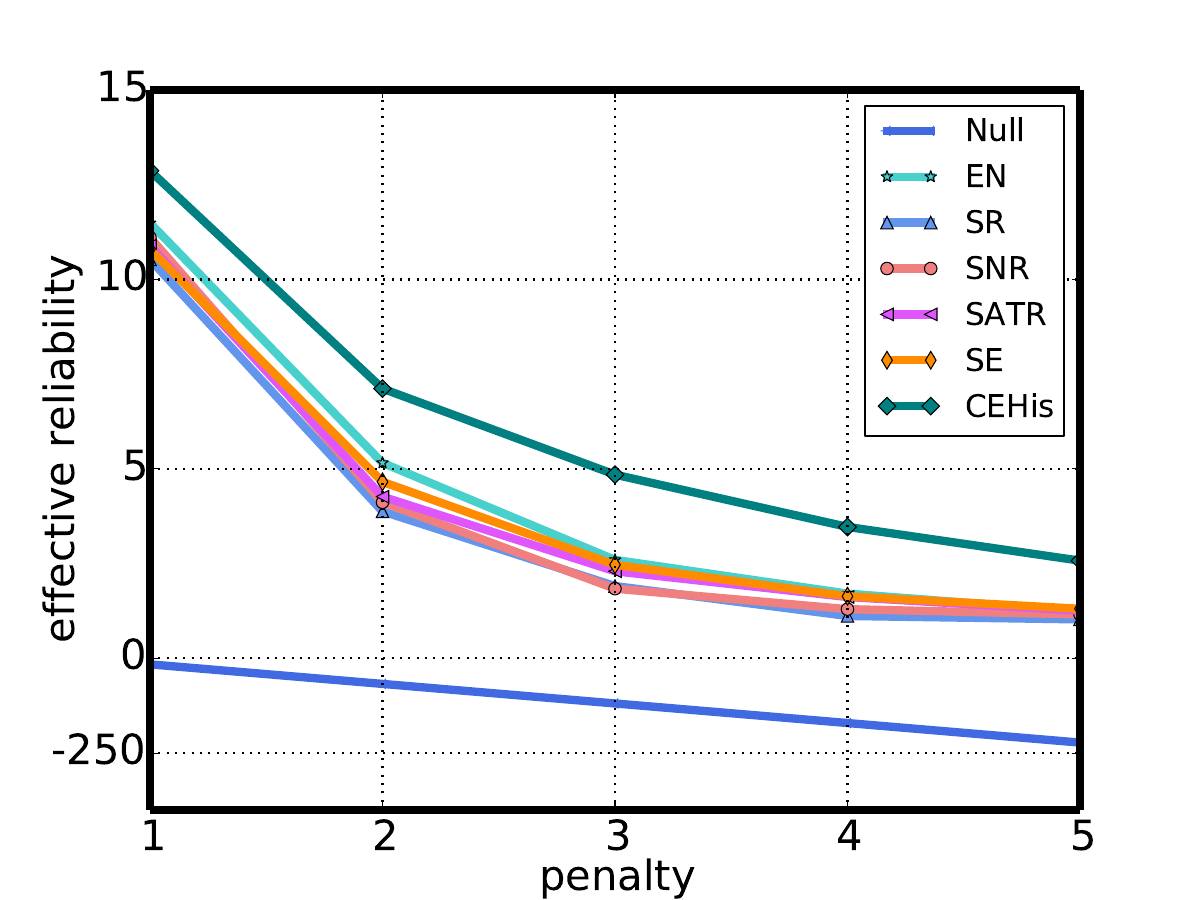}
  \label{fig:phi_renet_5}}
  \subfigure[RENET, N=10.] { 
  \includegraphics[width=0.465\linewidth]{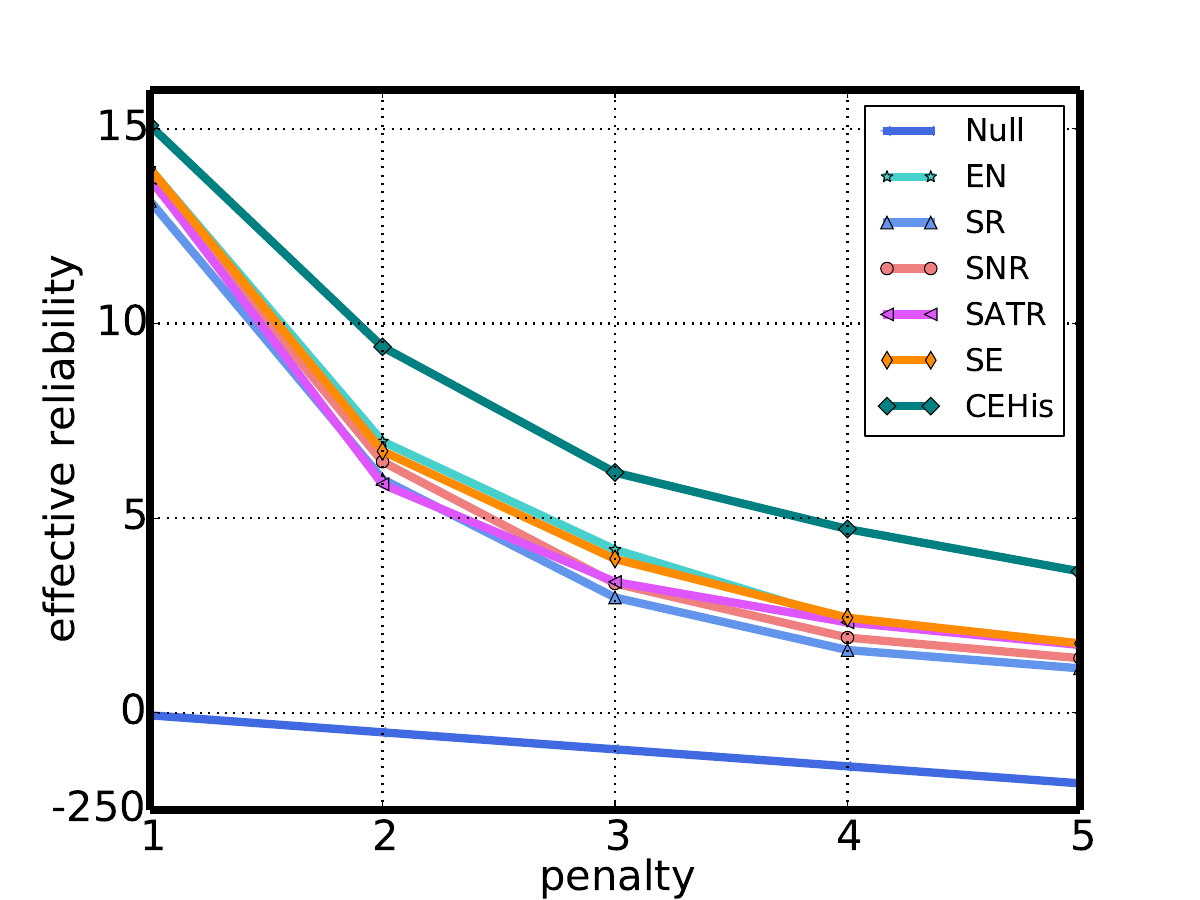}
  \label{fig:phi_renet_10}}\\
  \subfigure[REGCN, N=5.] { 
  \includegraphics[width=0.465\linewidth]{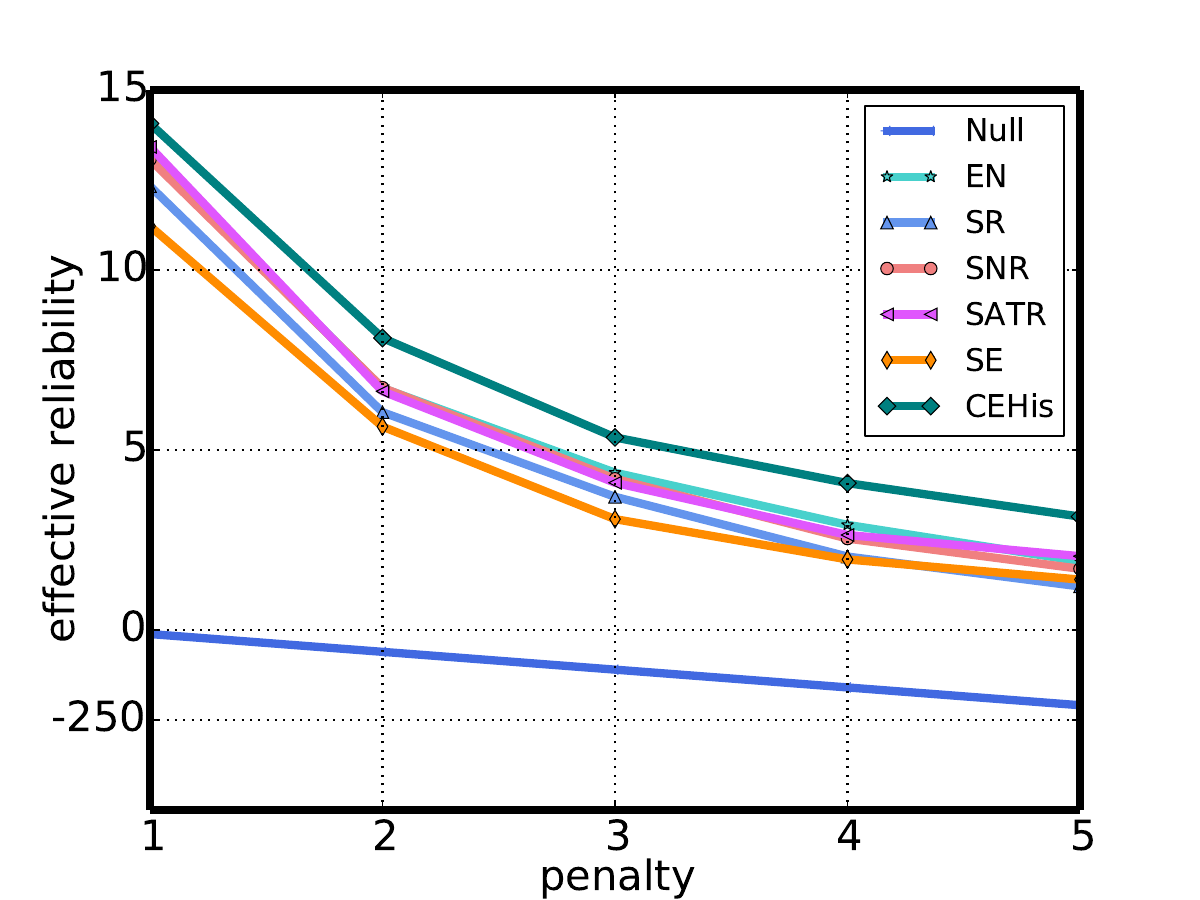}}
  \subfigure[REGCN, N=10.] { 
  \includegraphics[width=0.465\linewidth]{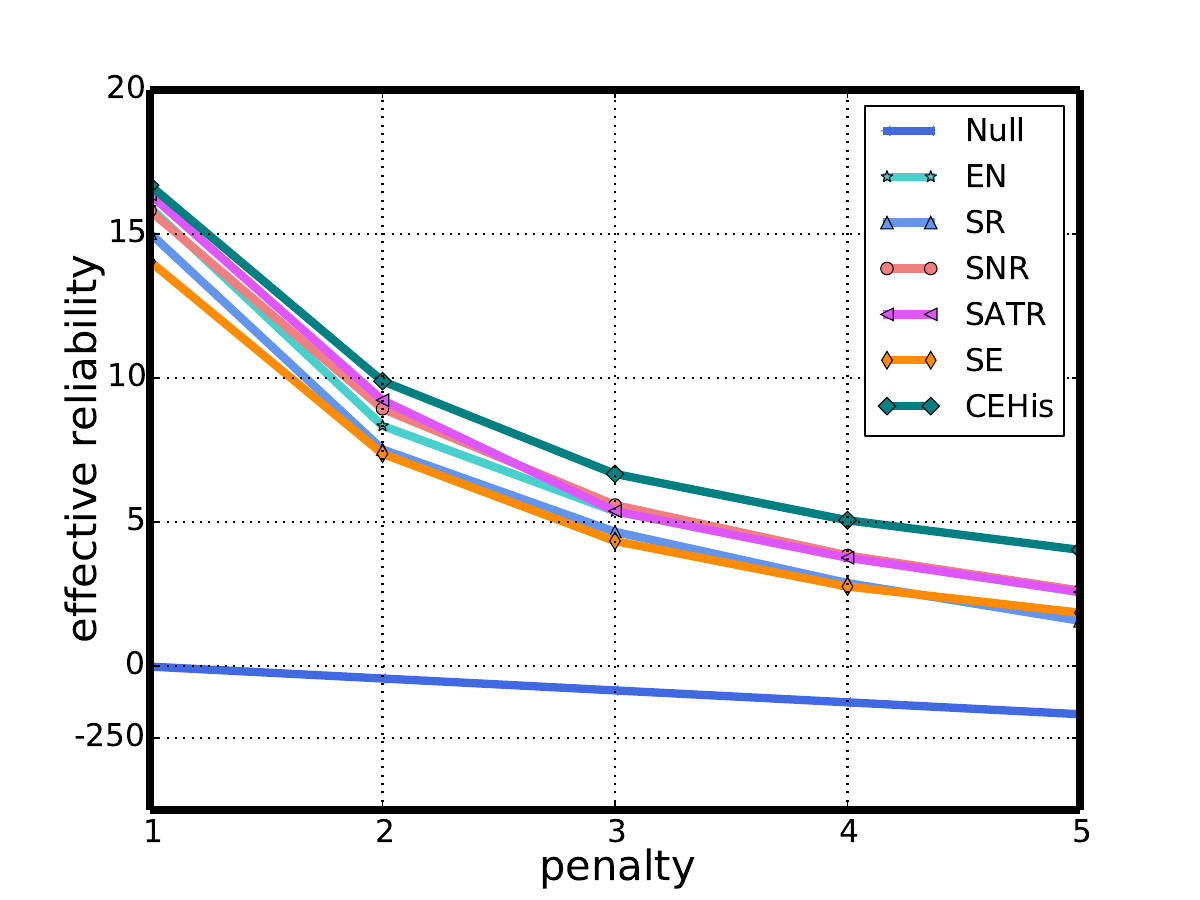}}\\
  \subfigure[TiRGN, N=5.] { 
  \includegraphics[width=0.465\linewidth]{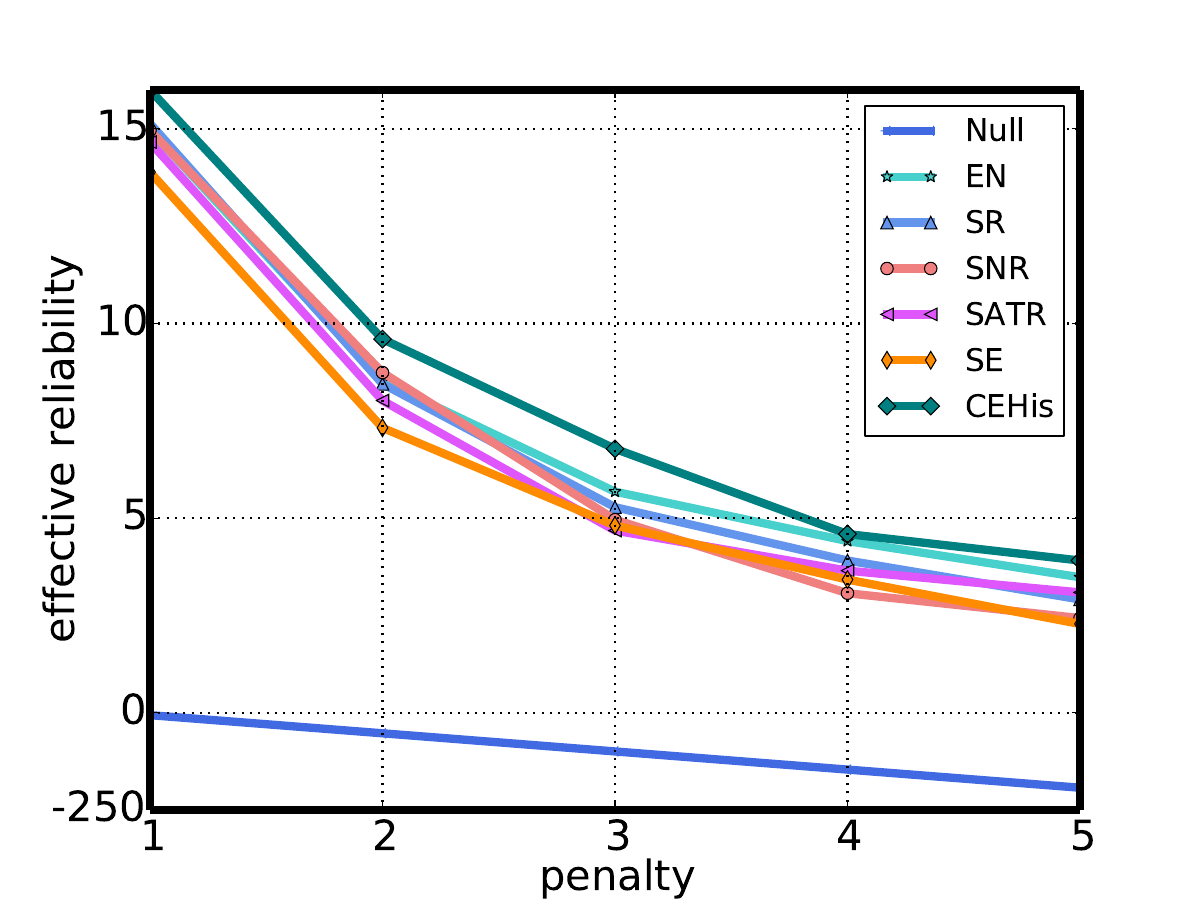}}
  \subfigure[TiRGN, N=10.] { 
  \includegraphics[width=0.465\linewidth]{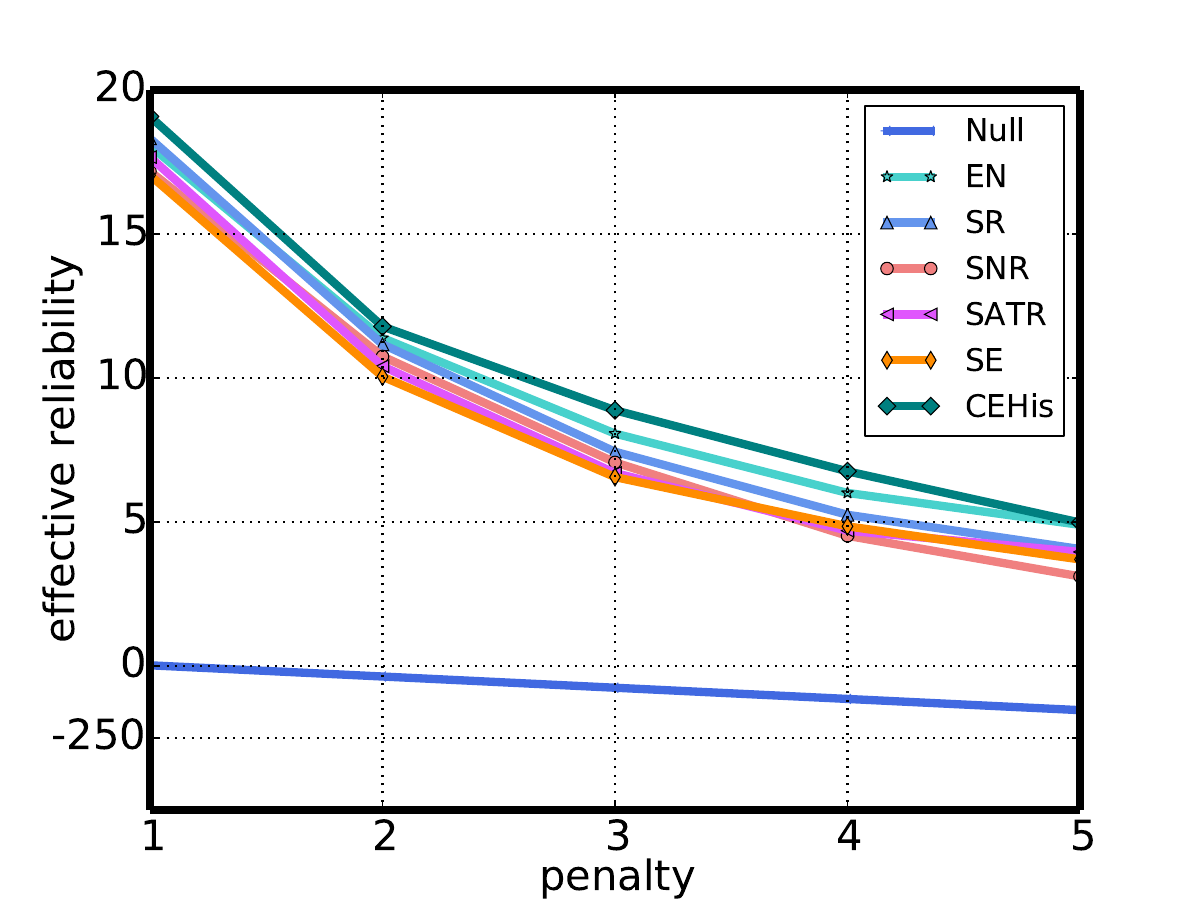}}
\caption{Effective reliability results of the selective entity reasoning task on ICEWS14. The penalty $c$ is set to 1, 2, 3, 4 and 5, while the model's tolerance $N$ is set to 5 and 10, respectively.}
  \label{fig:phi}
  \end{figure}

To examine the effectiveness of CEHis on the selective entity reasoning task, we focus on measuring the max coverage at different risk levels, the AUC for the risk-coverage curve, and the effective reliability scores under different penalty and tolerance settings. The results for the former two metrics are presented in Table~\ref{table:mainpred1}. Due to space limitation, for the latter metric, we only report the corresponding results on ICEWS14 in Figure~\ref{fig:phi}. 

From Table~\ref{table:mainpred1}, it can be observed that CEHis outperforms all baseline methods, in terms of AUC and the coverage under different risk levels in most cases. 
This is because all baseline methods do not model the accuracy of historical predictions on related queries. Since those related queries in the history are similar to the current one, their accuracy can serve as an indicator of the model's ability to make precise predictions for the current query. Therefore, by modeling the impact of the accuracy of historical predictions on the confidence of the current prediction, CEHis outperforms existing confidence estimators.

On ICEWS18, we notice that TiRGN with SR has lower coverage than REGCN with SR when the risk is 0.1 and 0.3. Considering that TiRGN has a higher parameter complexity of 13.98M (about 1.5x greater than that of REGCN), we guess TiRGN is overconfident in its incorrect predictions, which results in the above performance gap. The proposed CEHis alleviates the overconfidence problem in the complex TiRGN model, as the coverage of TiRGN with GEHis has been significantly improved when the risk level is  0.1 and 0.3.

Figure~\ref{fig:phi} shows that CEHis achieves the highest effective reliability scores across all penalty and tolerance levels. Also, it can be observed that when the penalty $c$ increases from 1 to 5, the effective reliability scores decrease quickly. For instance, in Figure~\ref{fig:phi_renet_5}, the RENET with CEHis has $\Phi_{1,5}>10$ and $\Phi_{2,5}<5$. It suggests that when the penalty for a wrong answer is high, the selective model will be more cautious and abstain from more predictions.

Additionally, we can observe that the performance of the selective TKG reasoning model is positively correlated with the model tolerance. For instance, in Figures~\ref{fig:phi_renet_5} and~\ref{fig:phi_renet_10}, the RENET with CEHis has $\Phi_{1,5}<15$ and $\Phi_{1,10}>15$.
A larger model tolerance $N$ allows the model to make more predictions, thus increasing the number of correct predictions. Furthermore, it can be noticed that the models without a confidence estimator consistently perform poorly when compared with their selective model counterparts. As the penalty increases, the performance gap between them becomes greater, which illustrates the necessity of empowering the TKG reasoning model with the ability to abstain from making predictions.

\begin{table}[ht]
  \setlength{\tabcolsep}{4pt}
  \centering
  \resizebox{0.95\linewidth}{!}{
  \begin{tabular}{c|cccccccc}
  \toprule
   \multicolumn{1}{c|}{\multirow{1}*{Model}}&EN &SR  &SNR  &SATR &SE &CEHis \\
  \midrule
  RENET  &43.37     &41.33     &41.79  &43.61   &43.89  &\textbf{40.23}  \\
    \midrule
  REGCN  &42.13     &39.91     &39.51  &39.09   &42.60  &\textbf{38.68}  \\
   \midrule
  TiRGN  &39.33     &36.83      &37.41  &36.79  &39.46  &\textbf{36.60}  \\
  \bottomrule
  \end{tabular}}
  \caption{AUC results of the selective relation reasoning task on ICEWS14. The risk ${R}$ is set to 0.1, 0.3 and 0.5.}
  \label{tab:rel-pre}
 % \vspace{-6mm}
  \end{table}

\subsubsection{Results on Selective Relation Reasoning}
To verify the effectiveness of CEHis on the selective relation reasoning task, we compare it with other confidence estimators.
Note that, for the selective relation reasoning task, CEHis focuses on the following three kinds of related queries, i.e., subject related, object related, and both subject and object related. Due to space limitation, we only report the AUC results on ICEWS14 in Table~\ref{tab:rel-pre}. We can see that CEHis performs better than baselines with different basic TKG reasoning models, which demonstrates again that modeling the accuracy of historical predictions is helpful for obtaining more accurate confidence scores in the selective TKG reasoning task.
\begin{figure} 
    \centering
  \subfigure[Results by different variants of CEHis with RENET.] {
  \includegraphics[width=\linewidth]{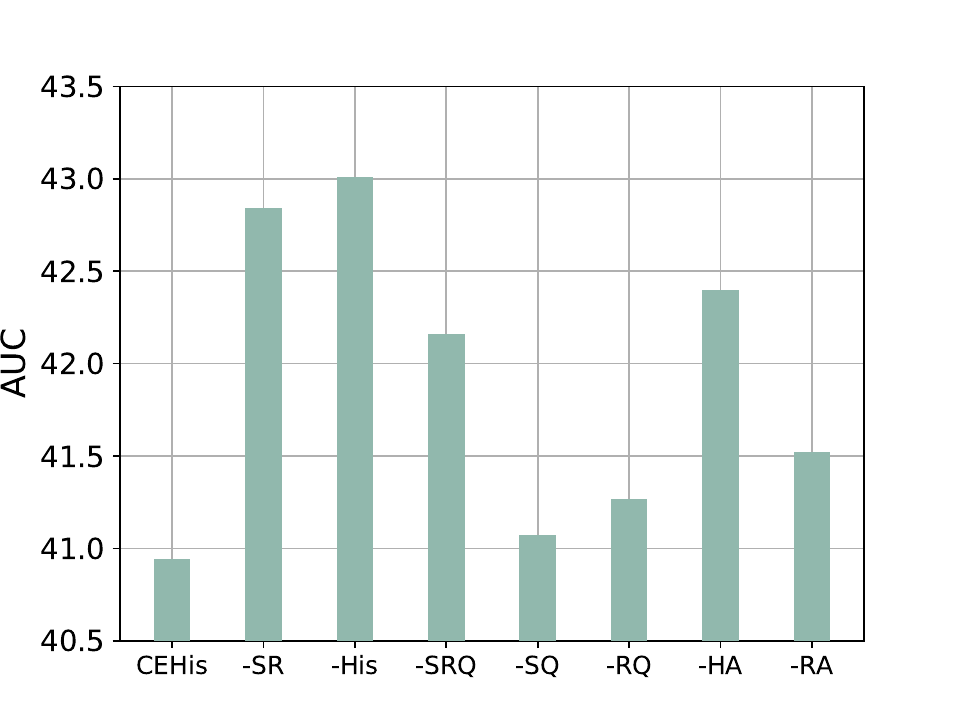}
  \label{fig:ablation_renet}}
  \subfigure[Results by different variants of CEHis with TiRGN.] { 
  \includegraphics[width=\linewidth]{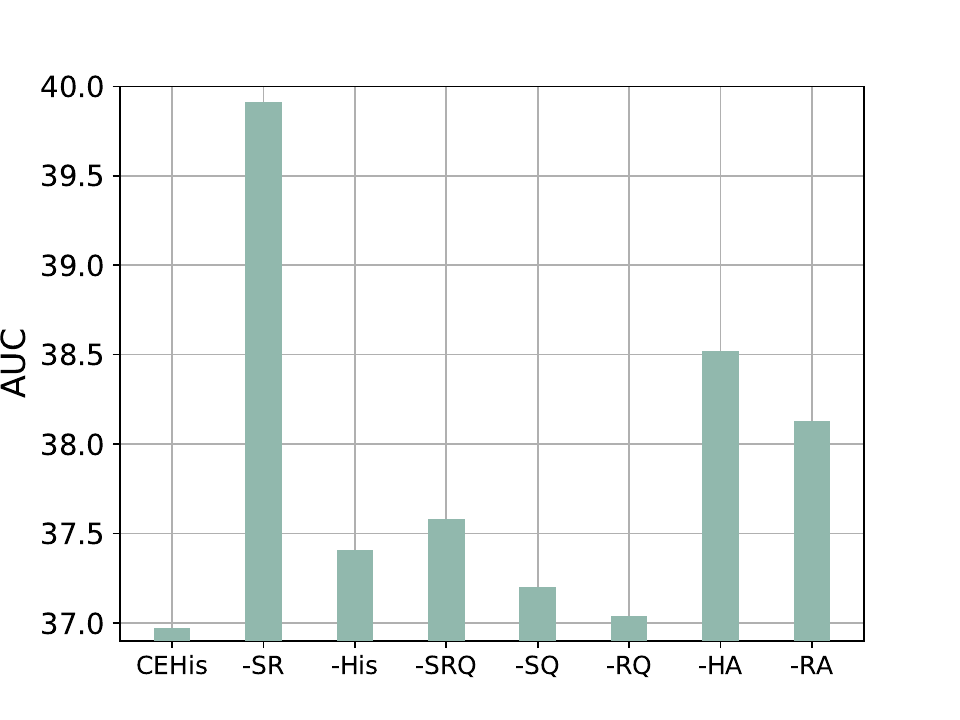}
  \label{fig:ablation_tirgn}}
  \caption{
  Comparison of variant models of CEHis
  with different basic TKG reasoning models on ICEWS14.}
  \label{fig:ablation}
\end{figure}

\begin{figure*}[!ht]
  \begin{center}
    \includegraphics[width=\linewidth]{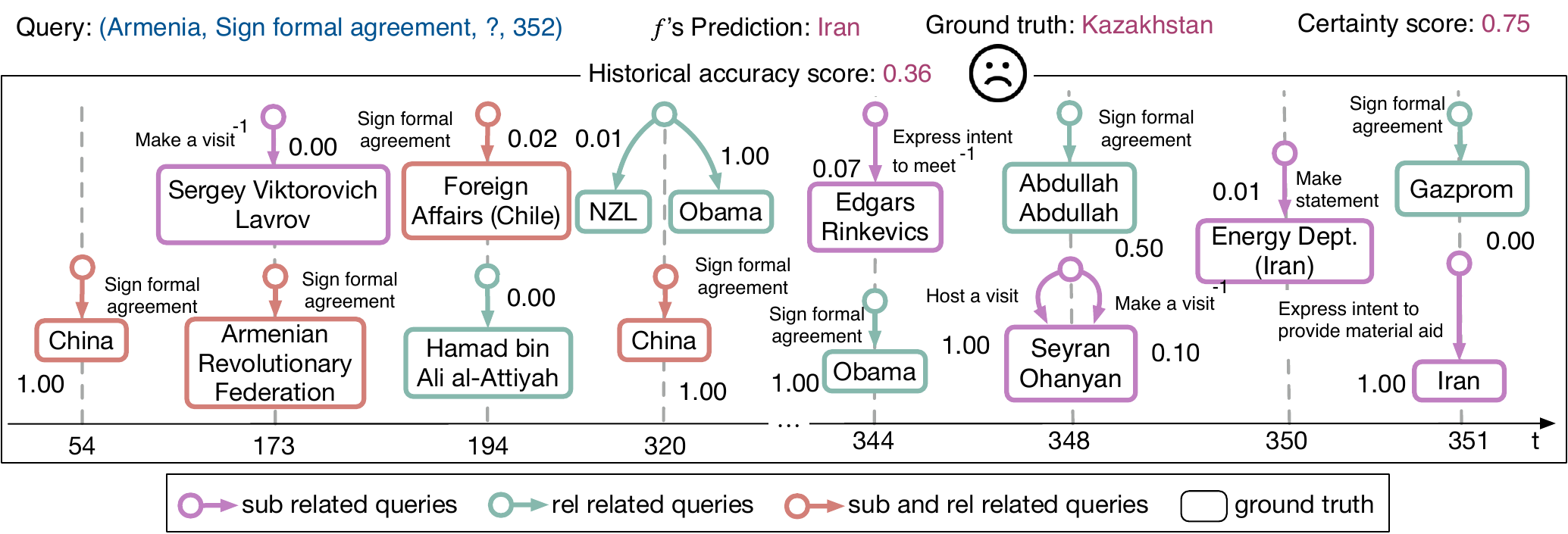}
  \caption{Case study on the necessity of modeling the accuracy of historical predictions. Each number represents how precise $f$'s historical prediction is on the corresponding query.}
  \label{fig:case}
  \end{center}
% \vspace{-6mm}
\end{figure*}

\subsection{Ablation Study}
To understand the behavior of CEHis with different basic TKG reasoning models on the selective TKG entity reasoning task, we take both RENET and TiRGN as the basic TKG reasoning models and conduct ablation studies on the ICEWS14 dataset. The corresponding AUC results are presented in Figure~\ref{fig:ablation}. 
In the following, we take RENET as an example, and analyze how each part of CEHis contributes to its performance with RENET.

From Figure~\ref{fig:ablation_renet}, we can observe that without considering the certainty of the current prediction (denoted as -SR) causes a drastic AUC increase, which indicates that the probability outputted by the basic TKG reasoning model is important, and should be taken into consideration when conducting the selective prediction task.

The result denoted as -His demonstrates the performance of CEHis without modeling the accuracy of historical predictions on related queries. It can be seen that, when employing RENET as the basic model, -His generates a higher AUC on ICEWS14, which justifies the necessity of modeling the accuracy of historical predictions.

To further analyze the importance of three kinds of related queries, we ignore the accuracy of historical predictions on both subject and relation related queries, subject related queries and relation related queries, denoted as -SRQ, -SQ and -RQ, respectively. As shown in Figure~\ref{fig:ablation_renet}, -SRQ brings the most significant performance drop when compared with -SQ and -RQ. This can be attributed to the fact that both subject and relation related queries contain the most useful information, which is also verified by existing TKG reasoning models~\cite{zhu2020learning}.

In addition, to verify the effectiveness of the Hawkes process (-HA in Figure~\ref{fig:ablation_renet}), we use the mean operation over the historical accuracy sequence of each kind of related queries. It can be seen that removing this part yields worse results compared to CEHis, demonstrating the necessity of utilizing the Hawkes process to model both the long-term and short-term impact of historical prediction accuracy.

To verify the effectiveness of the ranking-based aggregation (-RA in Figure~\ref{fig:ablation_renet}), we simply add the absolute value of the model's certainty of the current prediction and the accuracy of historical predictions on related queries. It can be seen that -RA results in worse performance compared to CEHis, demonstrating that the ranking-based strategy can help better aggregate the scores outputted by the certainty scorer and the historical accuracy scorer.

When taking TiRGN as the basic reasoning model, we can derive the same conclusion on different variants of CEHis, except for -His. From Figure~\ref{fig:ablation_renet} and Figure~\ref{fig:ablation_tirgn},  we observe that -His has a reduced influence on the final AUC result when employing TiRGN as the basic model compared to utilizing RENET. 
This is because TiRGN captures valuable information within repeated historical facts. This information can help TiRGN precisely predict queries with repeated ones in the history. As a result, the impact of the historical accuracy scorer on TiRGN is weakened.

\subsection{Case Study}
In order to further show the necessity of modeling the accuracy of historical predictions, we present a case study in Figure~\ref{fig:case} where the basic TKG reasoning model $f$ makes a wrong prediction. It can be observed that SR assigns a high confidence score (0.75) to the current prediction. However, the historical predictions on related queries are of low accuracy, which indicates that trusting the current prediction made by the basic TKG reasoning model may bring risk. CEHis utilizes the historical accuracy scorer to capture the accuracy of historical predictions, and guide the selective TKG reasoning model to abstain from making the current prediction. As a result, the risk brought by incorrect predictions can be controlled.

\section{Conclusions}
In this paper, we introduced the selection prediction setting for TKG reasoning, where a model is allowed to abstain in order to avoid making incorrect predictions. We further proposed a confidence estimator, called CEHis, to conduct the selective TKG reasoning task. CEHis considers both the certainty of the current prediction and the accuracy of historical predictions on related queries, and employs the Hawkes process to model the time-varying impact of the accuracy of historical predictions. Finally, we demonstrated the effectiveness of CEHis upon comparison with other confidence estimators by applying them to existing TKG reasoning models.
\section{Acknowledgments}
The work is supported by the National Natural Science Foundation of China under grant 62306299, the National Key Research and Development Project of China, Beijing Academy of Artificial Intelligence under grant BAAI2019ZD0306, the KGJ Project under grant JCKY2022130C039, and the Lenovo-CAS Joint Lab Youth Scientist Project. We thank anonymous reviewers for their insightful comments and suggestions.

\nocite{*}
\section{Bibliographical References}\label{sec:reference}
\bibliographystyle{lrec-coling2024-natbib}
\bibliography{lrec-coling2024-example}

\bibliographystylelanguageresource{lrec-coling2024-natbib}
\bibliographylanguageresource{languageresource}

\end{document}